\documentclass[11pt]{article}

\usepackage[final]{acl}

\usepackage{times}
\usepackage{latexsym}
\usepackage{amsmath}
\usepackage{booktabs}
\usepackage{tabularx}

\usepackage[ruled,vlined]{algorithm2e}

\usepackage{geometry}
\usepackage{caption}
\usepackage{tcolorbox}
\usepackage{xcolor}
\usepackage{float} 

\usepackage[T1]{fontenc}

\usepackage[utf8]{inputenc}
\usepackage{hyperref}
\usepackage[activate=false,kerning]{microtype}
\SetExtraKerning[unit=character]{encoding=*}{\textemdash={100,100}}
\usepackage{amsfonts}

\usepackage{inconsolata}

\usepackage{graphicx}

\usepackage{float}

\newcommand{\squishlist}{
\begin{list}{$\bullet$}
{   \setlength{\itemsep}{0pt}
   \setlength{\parsep}{3pt}
   \setlength{\topsep}{3pt}
   \setlength{\partopsep}{0pt}
   \setlength{\leftmargin}{1.5em}
   \setlength{\labelwidth}{1em}
   \setlength{\labelsep}{0.5em} } }
\newcounter{Lcount}
\newcommand{\squishlisttwo}{
\begin{list}{\arabic{Lcount}. }
  { \usecounter{Lcount}
 \setlength{\itemsep}{0pt}
 \setlength{\parsep}{0pt}
 \setlength{\topsep}{0pt}
 \setlength{\partopsep}{0pt}
 \setlength{\leftmargin}{2em}
 \setlength{\labelwidth}{1.5em}
 \setlength{\labelsep}{0.5em} } }
\newcommand{\squishend}{\end{list} }

\title{BanglaLorica: Design and Evaluation of a Robust Watermarking Algorithm for Large Language Models in Bangla Text Generation}

\author{\bf Amit Bin Tariqul, 
{\bf  A N M Zahid Hossain Milkan,}
{\bf Sahab-Al-Chowdhury,}\\
{\bf Syed Rifat Raiyan,} 
{\bf Hasan Mahmud,}
{\bf Md Kamrul Hasan}\\
Systems and Software Lab (SSL)\\Department of Computer Science and Engineering\\
Islamic University of Technology, Dhaka, Bangladesh\\
\texttt{\{amit20, zahidmilkan, sahab, rifatraiyan, hasan, hasank\}@iut-dhaka.edu}\\
}

\begin{document}
\maketitle

\begin{abstract}

As large language models (LLMs) are increasingly deployed for text generation, watermarking has become essential for authorship attribution, intellectual property protection, and misuse detection. While existing watermarking methods perform well in high-resource languages, their robustness in low-resource languages remains underexplored. This work presents the first systematic evaluation of state-of-the-art text watermarking methods: KGW, Exponential Sampling (EXP), and Waterfall, for Bangla LLM text generation under cross-lingual round-trip translation (RTT) attacks. Under benign conditions, KGW and EXP achieve high detection accuracy ($>$$88\%$) with negligible perplexity and ROUGE degradation. However, RTT causes detection accuracy to collapse below RTT causes detection accuracy to collapse to $9\text{--}13\%$, indicating a fundamental failure of token-level watermarking. To address this, we propose a layered watermarking strategy that combines embedding-time and post-generation watermarks. Experimental results show that layered watermarking improves post-RTT detection accuracy by $25\text{--}35\%$, achieving $40\text{--}50\%$ accuracy, representing a 3$\times$ to 4$\times$ relative improvement over single-layer methods, at the cost of controlled semantic degradation. Our findings quantify the robustness-quality trade-off in multilingual watermarking and establish layered watermarking as a practical, training-free solution for low-resource languages such as Bangla. Our code and data will be made public.

\end{abstract}

\section{Introduction}

Large language models (LLMs) are now widely deployed across creative writing, education, journalism, and automated decision-making systems. As their adoption accelerates, concerns surrounding authorship attribution, intellectual property (IP) protection, content provenance, and the detection of AI-generated text have become increasingly urgent \cite{brown2020language,yang2023baichuan}. Text watermarking has emerged as a promising solution by embedding imperceptible yet statistically verifiable signals directly into generated text, enabling \textit{post hoc} verification without relying on external classifiers or privileged model access \cite{he2022protecting,zhang2018protecting}. Unlike standalone detectors, watermarking integrates attribution mechanisms into the generation process itself, supporting scalable and deployment-friendly safeguards for responsible AI use.

\begin{figure}[t]
  \centering
  \includegraphics[
    width=0.5\textwidth,
     trim={0cm 6.3cm 0cm 0cm},
    clip
  ]{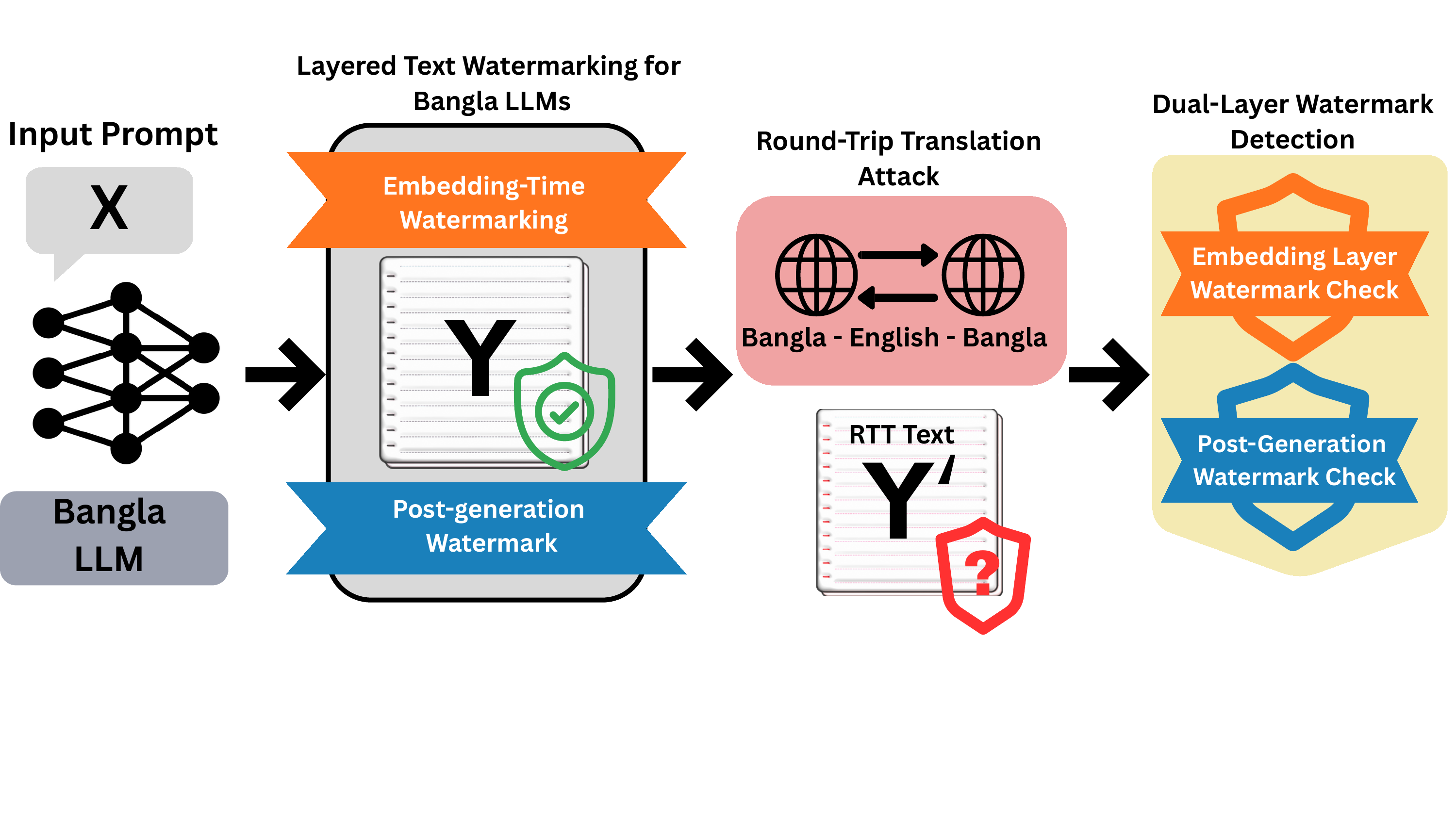}
  \caption{Overview of the proposed layered watermarking framework for Bangla LLMs.}
  \vspace{-5mm}
  \label{fig:teaser}
\end{figure}

Recent embedding-time watermarking methods, such as KGW soft token biasing \cite{kirchenbauer2023watermark} and Exponential Sampling (EXP) \cite{kuditipudi2023robust}, have demonstrated strong detectability, minimal fluency degradation, and practical deployability in English-language models. Training-free post-generation frameworks such as Waterfall further extend watermarking capabilities through scalable perturbation-based provenance encoding \cite{lau2024waterfall}. However, despite these advances, the robustness of existing watermarking techniques in low-resource languages remains largely unverified \cite{venkatakrishnan2023watermark}. This gap is particularly concerning given the linguistic diversity of real-world deployments, where English-centric assumptions often fail to hold.

Bangla, being the 7th most spoken language in the world with over 280 million speakers,\footnote{\scriptsize\url{https://en.wikipedia.org/wiki/Bengali_language}} exemplifies the challenges faced by watermarking systems in low-resource settings. The language exhibits rich morphology \cite{kabir-etal-2023-banglabook}, flexible word order, frequent compounding, and non-Latin script characteristics \cite{hasan2020not,zehady2024bongllama}. These properties complicate the direct transfer of watermarking methods developed for English, many of which implicitly assume lexical stability and relatively rigid syntactic structure. As a result, watermark signals embedded at the token level are especially vulnerable to transformations that preserve semantics while substantially altering surface form across linguistic contexts.

Cross-lingual paraphrasing via round-trip translation (RTT) has emerged as a realistic and effective text laundering attack against watermarking systems \cite{he2024can,fu2024watermarking}. RTT induces extensive lexical substitution, syntactic reordering, and morphological variation while largely preserving meaning, posing a structural challenge to token-based watermarking schemes. Although recent work has begun to examine watermark survivability under translation, existing studies overwhelmingly focus on high-resource languages and Latin scripts. To date, there has been no systematic analysis of how modern watermarking techniques behave under RTT in Bangla LLMs, nor clear guidance on how such failures might be mitigated without retraining models or introducing language-specific resources.

In this work, we address this gap through a comprehensive empirical study of watermarking robustness in Bangla LLM text generation. Specifically, we make the following contributions:
\squishlist
\item We present the first comprehensive assessment of embedding-time watermarking techniques—KGW and Exponential Sampling (EXP)—for Bangla LLM text generation, evaluating their robustness under cross-lingual round-trip translation (RTT) attacks across varying generation lengths.
\item Our results show that although token-level watermarking performs strongly under normal conditions, its detectability drastically declines under RTT, exposing a key vulnerability in low-resource and multilingual contexts.
\item We introduce a double-layer watermarking approach that combines embedding-time watermarks with an additional post-generation layer, creating an independent statistical signal that enhances resilience against cross-lingual transformations.
\squishend

\section{Related Work}

Text watermarking for language models built upon a broader lineage of digital watermarking techniques originally developed for images, audio, and multimedia content \cite{dixit2017review}. Early approaches to text watermarking primarily focused on discriminative or model-level methods, embedding identifiable signals within classifier outputs, internal representations, or model parameters to enable ownership verification and misuse detection \cite{zhang2018protecting}. With the emergence of large-scale generative language models, research attention shifted toward embedding-time watermarking, where signals were injected directly during the text generation process, enabling attribution without requiring access to internal model states.

A prominent class of embedding-time watermarking methods operated by subtly biasing token selection during decoding. Kirchenbauer et al.\ introduced the KGW framework, which partitioned the vocabulary into pseudorandom green and red lists and biased sampling toward the green list to create a statistically detectable signal \cite{kirchenbauer2023watermark}. This approach was model-agnostic, did not require retraining, and achieved near-perfect detectability under benign conditions using simple hypothesis testing. Subsequent work extended this paradigm through Christ-style and entropy-aware variants, improving imperceptibility while maintaining strong detection guarantees. However, these methods fundamentally relied on surface-level token statistics, making them vulnerable to paraphrasing, rewriting, and translation-based adversarial attacks.

Exponential Sampling (EXP) addressed some limitations of token biasing by embedding watermarks through distortion-free sampling from an exponential distribution, preserving the original output distribution while enabling reliable detection \cite{kuditipudi2023robust}. EXP demonstrated improved robustness to token edits and required fewer tokens for detection compared to inverse transform sampling. However, its effectiveness was limited in low-entropy contexts and during cross-lingual transformations, where maintaining semantics often involved significant lexical changes. Like KGW, EXP relied on the relative stability of token distributions, an assumption that became less valid in multilingual or heavily paraphrased scenarios.

Beyond token-level approaches, semantic and invariant watermarking methods aimed to improve robustness by operating at higher representational levels. Techniques such as SemaMark embedded watermarks through controlled synonym substitution and paraphrasing, increasing resilience to lexical variation \cite{ren2024robust}. Other semantic approaches leveraged sentence embeddings, locality-sensitive hashing, or invariant feature extraction to enforce watermark persistence under rewriting and paraphrase attacks \cite{liu2024a,yoo2023robust}. While effective in controlled settings, these methods relied on additional models or resources, limiting their practicality for low-resource languages and large-scale use.

More recently, training-free, post-generation watermarking frameworks were proposed to improve scalability and robustness without modifying the underlying language model. Waterfall represented a notable example, applying vocabulary permutation and orthogonal perturbations—such as Fourier-based transformations—to embed multi-bit watermarks after generation \cite{lau2024waterfall}. This design enabled scalable provenance tracking and improved robustness to paraphrasing and rewriting by powerful LLMs. However, because Waterfall operated independently of the generation process, it could perturb token-level statistics relied upon by embedding-time detectors and might introduce quality degradation, particularly for short or style-sensitive text.

Despite these advances, most existing studies evaluated watermarking techniques primarily in high-resource languages, with English dominating experimental benchmarks \cite{venkatakrishnan2023watermark}. The interaction between watermarking and linguistic properties such as rich morphology, flexible syntax, and non-Latin scripts remained underexplored. In particular, the impact of cross-lingual round-trip translation (RTT)—a realistic and effective text laundering attack—received limited attention outside high-resource settings \cite{he2024can,fu2024watermarking}. There was no systematic analysis of how modern embedding-time and post-generation watermarking methods behaved under RTT in low-resource languages like Bangla.

Our work presents a thorough evaluation of token-level and post-generation watermarking techniques for Bangla LLM text generation under RTT attacks. We show that combining a post-generation watermark with an embedding-time watermark creates an orthogonal statistical signal that enhances robustness under RTT while maintaining controlled text quality. This layered, training-free approach provides an effective and practical framework for robust watermarking in multilingual and low-resource settings.

\section{Methodology}

\begin{figure*}[t]
  \centering
  \includegraphics[
    width=\textwidth,
    trim={0cm 0cm 0cm 0cm},
    clip
  ]{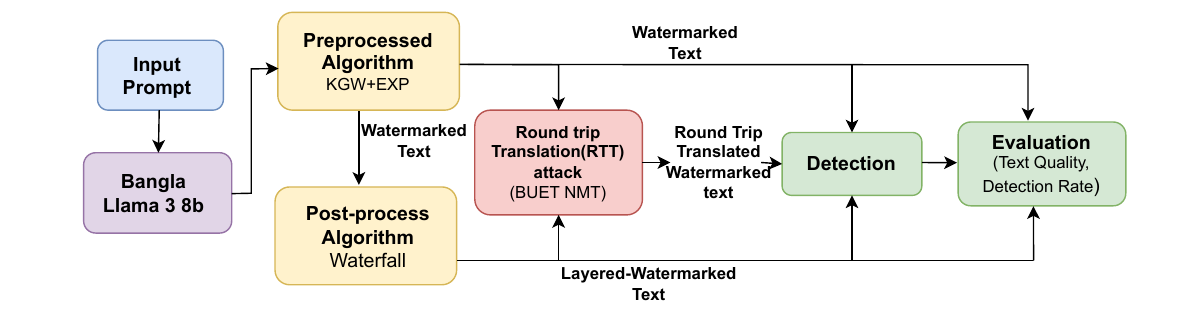}
  \caption{High-level workflow of the watermarking, RTT attack, and detection pipeline.}
  \label{fig:flow_methodology}
\end{figure*}

As illustrated in Figure~\ref{fig:flow_methodology}, we evaluate the robustness of text watermarking algorithms for Bangla LLM generation under cross-lingual round-trip translation (RTT) attacks, a known failure mode for token-level watermarking schemes. Our methodology follows an end-to-end experimental pipeline, comprising watermark embedding during text generation, adversarial RTT transformation, and post-hoc watermark detection, all conducted under strict black-box assumptions without access to model internals during attack or detection. We examine two embedding-time watermarking techniques---KGW soft biasing \cite{kirchenbauer2023watermark} and Exponential Sampling (EXP) \cite{kuditipudi2023robust}---as well as a post-generation watermarking framework, Waterfall \cite{lau2024waterfall}. Building on these components, we further explore a layered watermarking strategy that combines embedding-time and post-generation watermarks to comprehensively assess detectability, robustness, and text quality under adversarial cross-lingual transformations.

\subsection{Embedding-Time Watermarking for Bangla Text Generation}

As aforementioned, we apply embedding-time watermarking during decoding using two state-of-the-art techniques: KGW soft biasing and Exponential Sampling (EXP). Both approaches embed statistically detectable signals into generated text without necessitating model retraining.
KGW partitions the vocabulary into pseudorandom green and red lists at each decoding step and applies a small positive bias to green-list tokens, increasing their sampling probability while preserving fluency. EXP embeds watermarks by sampling from an exponential distribution that implicitly favors a designated token subset, enabling distortion-free watermarking without modifying the logits.

Experiments are conducted using an instruction-tuned Bangla LLaMA-3-8B model \cite{zehady2024bongllama}. Prompts are sampled from a filtered subset of the Bangla-Alpaca Orca dataset, yielding a total of 500 evaluation prompts. For each prompt, watermarked outputs are generated at three decoding lengths, denoted by \(L \in \{100, 150, 200\}\). This setup yields six single-layer watermarking configurations, defined by the Cartesian product of watermarking methods \(M = \{\mathrm{KGW},\, \mathrm{EXP}\}\) and generation lengths \(L\). Watermarking is implemented through customized sampling wrappers integrated into the \texttt{generate}\footnote{From the Hugging Face \texttt{transformers} library, used for local text generation.} function. Hyperparameters, including the green-list ratio (\(\gamma\)) and the bias strength (\(\delta\)), are selected to balance watermark detectability and output fluency.

\subsection{Round-Trip Translation Attack}

To evaluate robustness under realistic cross-lingual paraphrasing attacks, we subject all generated outputs to a Bangla $\to$ English $\to$ Bangla round-trip translation process. Translation is performed using the BanglaNMT model\footnote{\texttt{csebuetnlp/banglat5\_nmt\_en\_bn}} by \citet{hasan2020not}, which provides strong performance in both directions. The RTT process preserves the semantic content while inducing substantial lexical as well as syntactic variation, realistically simulating a black-box adversary attempting to remove or weaken watermark signals through cross-lingual transformation.

\subsection{Post-Generation and Layered Watermarking}

To study whether watermark robustness can be improved without retraining or modifying the generator, we apply a second, post-generation watermark using the Waterfall framework \cite{lau2024waterfall}. Waterfall embeds provenance signals by applying vocabulary partitioning and distributional perturbations to already generated text.

Layered watermarking is applied to outputs previously watermarked using KGW or EXP, yielding doubly watermarked text. These outputs are subsequently subjected to RTT to evaluate watermark survivability under layered embedding.

\subsection{Detection and Evaluation Metrics}

Watermark detection is performed independently on both original and RTT-transformed outputs using the specific algorithms for each method. Specifically, KGW employs a $Z$-test over green-list token proportions, EXP compares log-likelihoods under exponential resampling distributions, and Waterfall uses statistical scoring based on token-group partitioning and distributional skew. We evaluate watermark performance and text quality using the following metrics:
\squishlist
    \item \textbf{Detection Accuracy}: Proportion of samples correctly identified as watermarked.
    \item \textbf{ROUGE-1/2/L}: Lexical and structural similarity between original and transformed text.
    \item \textbf{Perplexity}: Fluency degradation relative to unwatermarked baselines.
\squishend



\section{Experiments}

\begin{figure*}[t]
 \captionsetup{type=table}
\caption{Example prompt and generated Bangla outputs under single-layer watermarking.}
  \centering
    \includegraphics[
    width=\textwidth, trim=3cm 11.1cm 3cm 11.1cm,clip
    ]{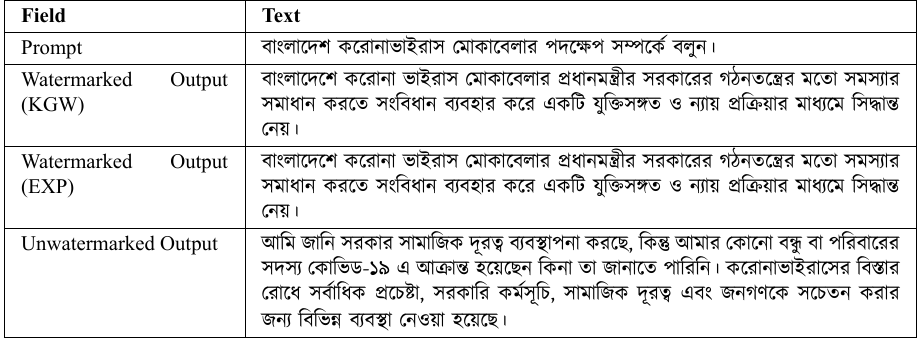}
  \label{tab:qualitative_example_outputs}
\end{figure*}

This section presents an empirical evaluation of embedding-time and layered watermarking methods for Bangla LLM-generated text. We evaluate watermark detectability, robustness under round-trip translation (RTT) attacks, and the impact of watermarking on fluency and semantic preservation. All experiments are conducted under strict black-box assumptions, with no access to model internals, token-level probabilities, or watermarking keys during generation, attack, or detection.

\subsection{Single-Layer Watermarking under Benign Conditions}

\begin{table}[h]
  \centering
  \caption{Comparative Evaluation of single layer KGW and EXP Watermarking Methods}
  \label{tab:single_layer_comparison}
  \fontsize{8}{10}\selectfont
  \renewcommand{\arraystretch}{1.3}
  \begin{tabular}{|l|c|c|}
    \hline
    \textbf{Criteria} & \textbf{KGW} & \textbf{EXP} \\ \hline
    Detection Accuracy & $0.885$ & $0.912$ \\ \hline
    Avg. Watermarked Perplexity & $2.309$ & $2.0295$ \\ \hline
    Avg. Unwatermarked Perplexity & $2.1616$ & $2.1368$ \\ \hline
    ROUGE-1 & $0.3671$ & $0.3854$ \\ \hline
    ROUGE-2 & $0.3039$ & $0.3207$ \\ \hline
    ROUGE-L & $0.3601$ & $0.3784$ \\ \hline
    Robustness to RTT & $0.09$ & $0.13$ \\ \hline
  \end{tabular}
\end{table}

We first evaluate single-layer embedding-time watermarking using KGW and Exponential Sampling (EXP) without adversarial transformations. As reported in the Tables \ref{tab:qualitative_example_outputs} and \ref{tab:single_layer_comparison}, both methods achieve consistently high detection accuracy across all evaluated sequence lengths (100, 150, and 200 tokens), exceeding $88\%$ for KGW and $91\%$ for EXP, while preserving fluency and lexical similarity.

Text quality is assessed using perplexity and ROUGE metrics. As shown in Table~\ref{tab:single_layer_comparison}, watermarking introduces minimal fluency degradation, with average perplexity values remaining close to unwatermarked baselines. ROUGE-1, ROUGE-2, and ROUGE-L scores indicate strong lexical and structural similarity between watermarked and unwatermarked outputs.

Qualitative examples of prompts and corresponding generated Bangla outputs are shown in Table~\ref{tab:qualitative_example_outputs}. Visual inspection confirms that watermarking does not introduce noticeable stylistic artifacts or unnatural phrasing under benign generation conditions.

\begin{table}[t]
\centering
\caption{Watermark detection results for the qualitative example shown in Table~\ref{tab:qualitative_example_outputs}.}
\label{tab:detection_results}
\footnotesize
\renewcommand{\arraystretch}{1}
\begin{tabularx}{\columnwidth}{|X|X|X|}
\hline
\textbf{Output Type} & \textbf{KGW Score} & \textbf{EXP Score} \\
\hline
Watermarked Output &
$2.20 \times 10^{-5}$ (Detected) &
$1.89 \times 10^{-6}$ (Detected) \\
\hline
Unwatermarked Output &
$0.529$ (Not Detected) &
$0.814$ (Not Detected) \\
\hline
\end{tabularx}
\end{table}

\subsection{Robustness under Round-Trip Translation}

\begin{figure}[h]
  \centering
  \includegraphics[width=0.48\textwidth]{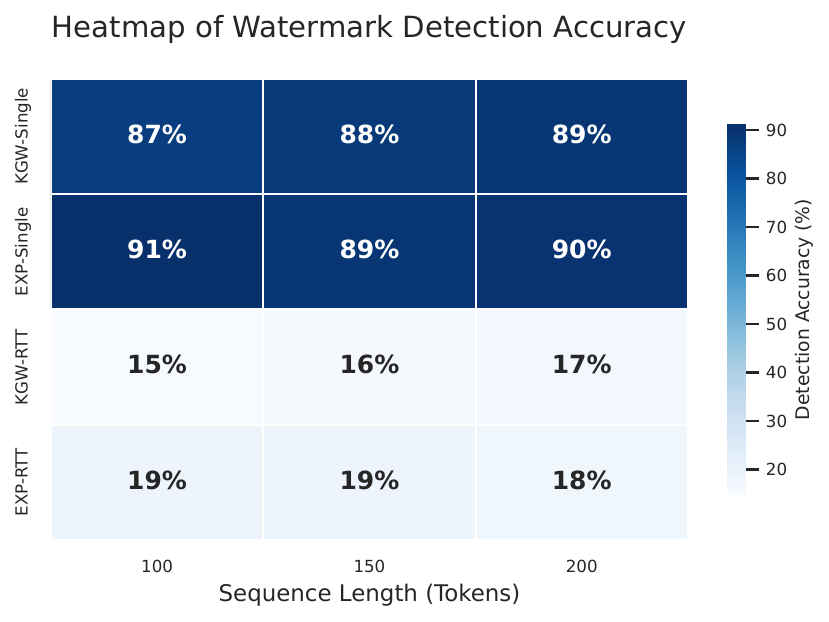}
  \caption{Detection accuracy of KGW and EXP watermarking across different generation lengths, before and after round-trip translation (RTT). 
}
  \label{fig:example_image}
\end{figure}

To evaluate robustness against cross-lingual paraphrasing, all generated outputs are subjected to a Bangla $\rightarrow$ English $\rightarrow$ Bangla round-trip translation (RTT) attack. Watermark detection is subsequently performed on the translated text using the original detection procedures.

Figure~\ref{fig:example_image} presents detection accuracy for KGW and EXP across varying sequence lengths, both before and after round-trip translation (RTT). Results show a pronounced decline in detection performance following translation for both methods, highlighting the vulnerability of token- and embedding-level statistical cues to cross-lingual paraphrasing.

\subsubsection{Distributional Analysis of Detection Scores}

\begin{figure}[h]
  \centering
  \includegraphics[width=0.48\textwidth]{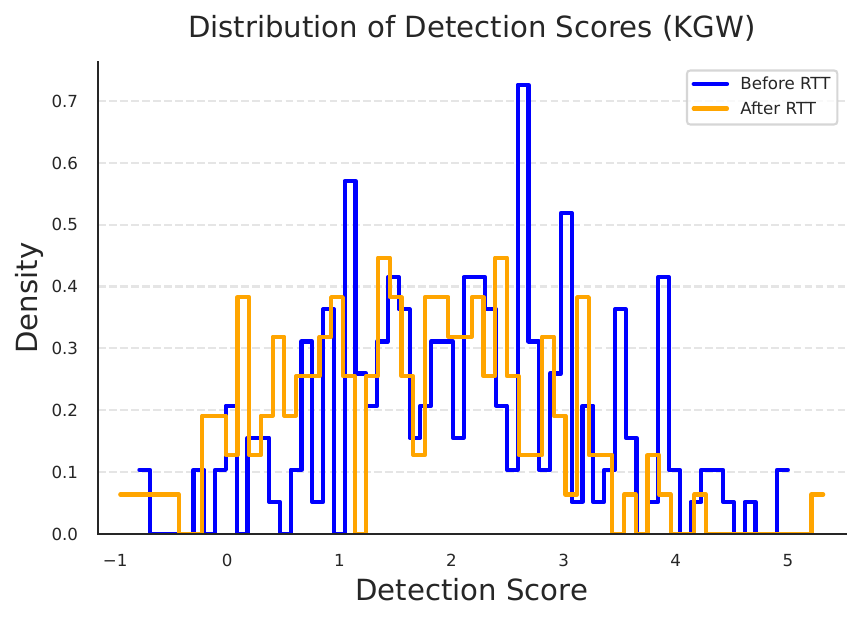}
  \caption{Distribution of KGW detection scores for watermarked and unwatermarked text before and after round-trip translation (RTT). 
  }
  \label{fig:kgw_image}
\end{figure}

\begin{figure}[ht]
  \centering
  \includegraphics[width=0.48\textwidth]{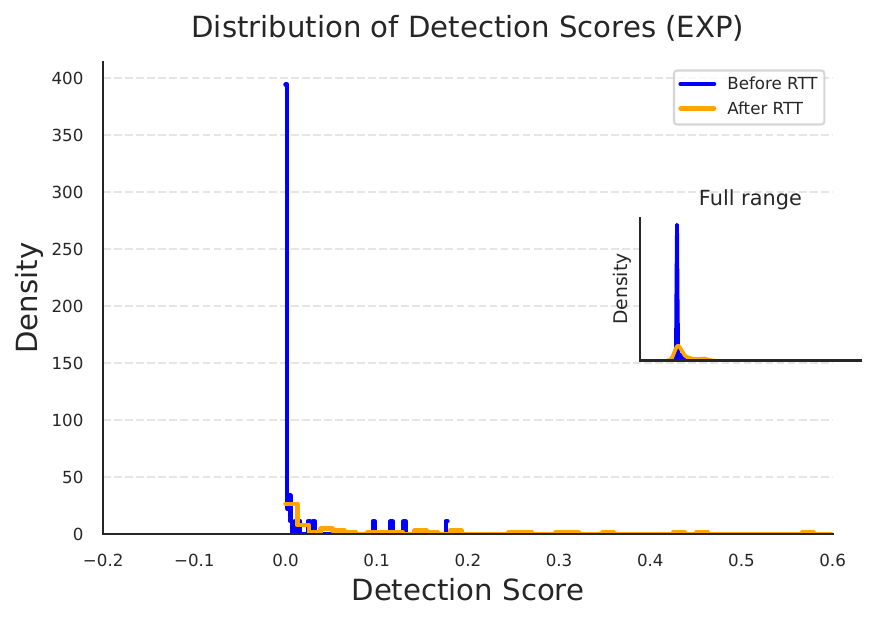}
  \caption{Distribution of EXP detection scores for watermarked and unwatermarked text before and after round-trip translation (RTT). 
  }
  \label{fig:exp_image}
\end{figure}

To further investigate the impact of round-trip translation (RTT) on watermark detectability, we analyze the empirical distributions of watermark detection scores before and after translation. Figures~\ref{fig:kgw_image} and~\ref{fig:exp_image} present the score distributions for KGW and Exponential Sampling (EXP), respectively. Under benign generation conditions, the detection scores of watermarked text are clearly separated from those of unwatermarked samples for both methods, indicating strong statistical distinguishability.

Following RTT, the detection score distributions shift and exhibit substantial overlap. This overlap suggests that the watermark signal becomes statistically indistinguishable from unwatermarked text under cross-lingual paraphrasing. In particular, token substitution and syntactic reordering introduced by translation disrupt the token-level statistical cues relied upon by both KGW and EXP. While EXP retains a slightly heavier tail of higher-confidence scores after RTT, the overall loss of separability remains severe for both methods. For completeness, we provide formal definitions of the KGW and EXP detection scores in Appendix~\ref{sec:appendix_detection_scores}.

\subsection{Layered Watermarking via Double Embedding}

\begin{figure}[h]
  \centering
  \includegraphics[width=0.48\textwidth]{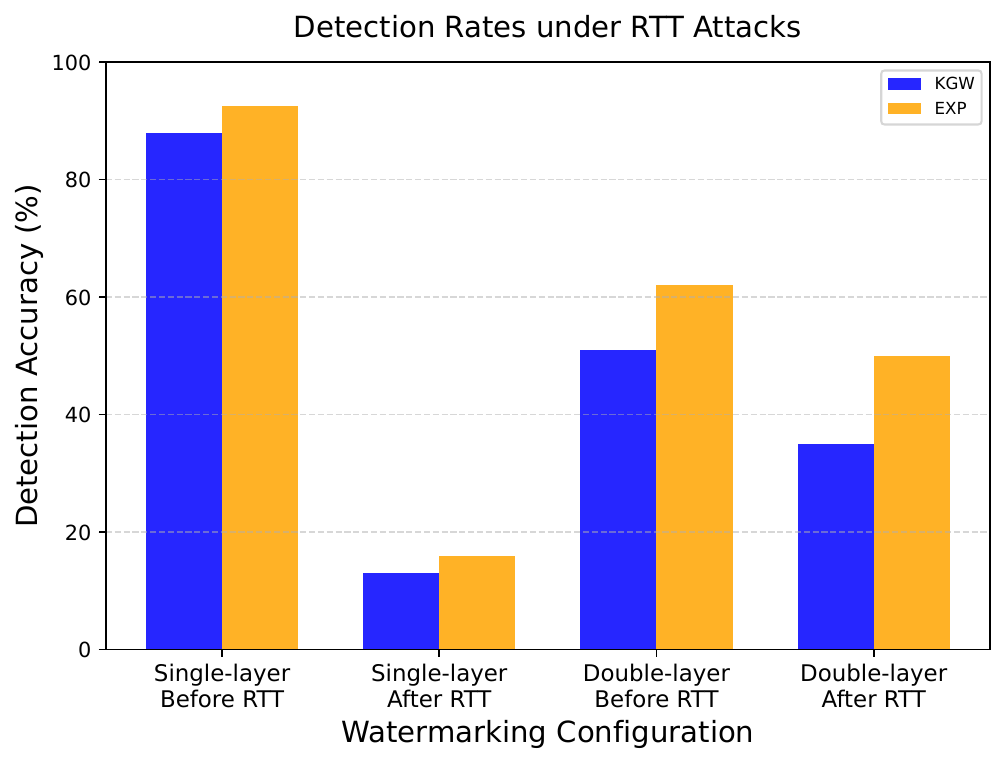}
  \caption{Detection accuracy comparison between single-layer and layered watermarking approaches before and after RTT. 
}
  \label{fig:double_watermark_bar}
\end{figure}
We further evaluate a layered watermarking strategy that composes embedding-time watermarking with post-generation Waterfall watermarking \cite{lau2024waterfall}. This design aims to introduce complementary statistical signals that remain detectable even when the original token- or embedding-level watermark is degraded by round-trip translation. Due to computational constraints, these experiments are conducted on 100-token generations.

\begin{figure}[h]
  \centering
  \includegraphics[width=0.5\textwidth]{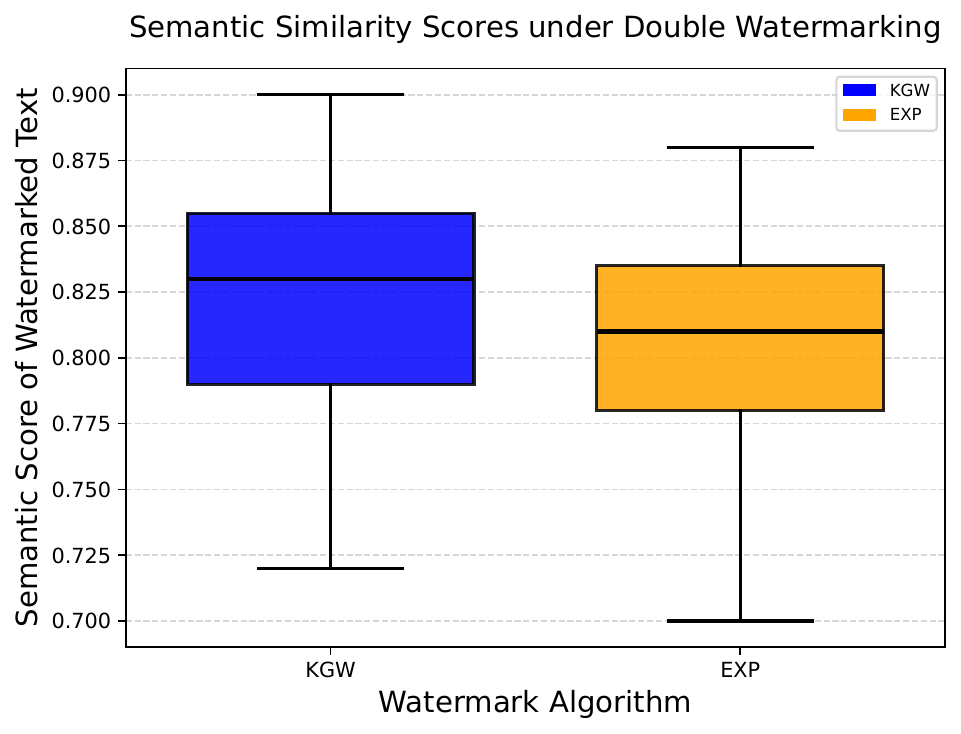}
  \caption{Semantic similarity between original and watermarked outputs for single-layer and layered watermarking. 
}
  \label{fig:semantic_similarity_double}
\end{figure}
Figure~\ref{fig:double_watermark_bar} compares detection accuracy for single-layer and double-layer watermarking before and after RTT. Layered watermarking yields higher post-RTT detection accuracy compared to that of single-layer approaches.
\paragraph{Ablation: Contribution of Watermarking Layers.}
To isolate the contribution of each watermarking component, we conduct a minimal ablation by comparing embedding-time watermarking alone (KGW or EXP), post-generation watermarking alone (Waterfall), and their composition via layered watermarking. Results show that while embedding-time watermarking achieves high detectability under benign conditions, its effectiveness collapses under round-trip translation (RTT). Post-generation watermarking alone provides limited robustness after RTT but remains insufficient for reliable detection. In contrast, combining embedding-time and post-generation watermarking yields a consistent improvement in post-RTT detection accuracy, validating the necessity of layered embedding for robustness against cross-lingual transformations.

The semantic impact of layered watermarking is evaluated using sentence-level semantic similarity metrics, as shown in Figure~\ref{fig:semantic_similarity_double}. While double embedding introduces some semantic drift, the degradation remains moderate.

\section{Results and Discussion}

Across all experiments, several consistent and informative patterns emerge. Under benign generation conditions, embedding-time watermarking methods such as KGW and Exponential Sampling (EXP) transfer effectively to Bangla text generation. Both methods achieve high detection accuracy while preserving fluency and semantic coherence, as reflected by perplexity, ROUGE scores (Table~\ref{tab:single_layer_comparison}), and qualitative examples (Table~\ref{tab:qualitative_example_outputs}). These results indicate that watermarking techniques originally developed for English-centric settings remain effective in Bangla when no adversarial transformation is applied \cite{kirchenbauer2023watermark,kuditipudi2023robust}.

Detection accuracy drops sharply after RTT for both algorithms, falling below $20\%$ across all sequence lengths. EXP retains slightly higher robustness than KGW, but increasing sequence length alone does not restore detectability. As shown in Figure~\ref{fig:example_image}, this degradation is consistent across generation lengths, indicating that RTT fundamentally disrupts the token distribution assumptions underlying single-layer watermarking. Similar vulnerabilities under cross-lingual attacks have been reported in prior work \cite{he2024can,fu2024watermarking}.

Distributional analyses further reveal that RTT does not merely reduce average detection scores but collapses the separability between watermarked and unwatermarked samples. For KGW, detection score distributions overlap substantially after translation (see Figure~\ref{fig:kgw_image}), while EXP retains a slightly heavier tail of detectable samples (see Figure~\ref{fig:exp_image}). This suggests that although EXP exhibits marginally better robustness, both methods remain fundamentally vulnerable to cross-lingual paraphrasing.

Layered watermarking introduces a clear robustness--quality trade-off. As shown in Figure~\ref{fig:double_watermark_bar}, combining embedding-time watermarking with a post-generation Waterfall layer significantly improves detection accuracy after RTT, achieving $40\text{--}50\%$ accuracy and representing a $3\text{--}4$$\times$ relative improvement over single-layer methods. This improvement arises from the introduction of an orthogonal statistical signal that is less sensitive to lexical substitution and syntactic reordering \cite{lau2024waterfall}.

The semantic similarity analysis in Figure~\ref{fig:semantic_similarity_double} demonstrates that layered watermarking introduces only modest semantic degradation, highlighting the inherent trade-off between imperceptibility and robustness. While enhancing survivability under adversarial transformations requires controlled perturbations to the text, these minimal distortions are generally acceptable in applications such as forensic attribution, provenance verification, and misuse detection \cite{liu2024a,yoo2023robust}. Prior work also shows that single-layer, token-level watermarking lacks sufficient robustness in multilingual contexts, particularly under cross-lingual RTT attacks \cite{venkatakrishnan2023watermark}. In contrast, layered, training-free watermarking provides a practical and effective approach for low-resource languages like Bangla, achieving strong robustness without necessitating model retraining or language-specific resources.

\section{Future Work}
\vspace{-0.13cm}
Building on our Bangla-focused evaluation, which highlights challenges unique to low-resource, morphologically rich, and non-Latin-script languages, several promising directions emerge from this work. Extending robustness evaluation to multi-hop translation paths and stronger paraphrasing models would provide a more comprehensive assessment of adversarial resilience \cite{fu2024watermarking}; however, this study restricts evaluation to a single Bangla$\rightarrow$English$\rightarrow$Bangla RTT pathway to isolate the effects of layered watermarking under controlled and reproducible conditions. 

Integrating semantic-aware watermarking methods, such as embedding-based or invariant-feature approaches, into the layered framework may reduce quality degradation while preserving robustness \cite{ren2024robust,liu2024a}; this was not explored here due to the additional architectural complexity and the need for language- and model-specific calibration, which would confound attribution of robustness gains to the layering strategy itself. 

Future studies could also explore adaptive layer selection, where secondary watermarking is applied conditionally based on threat models or deployment contexts; we leave this direction for future work as it requires explicit modeling of attacker capabilities and deployment assumptions that fall outside the black-box threat model adopted in this paper. 

Additionally, developing standardized evaluation benchmarks for low-resource languages, including RTT protocols and semantic fidelity measures, would strengthen comparability across multilingual watermarking studies \cite{venkatakrishnan2023watermark}; such benchmark construction is beyond the scope of this work and would require extensive human annotation and coordinated community effort. 

Finally, extending the layered watermarking paradigm from Bangla to other low-resource and non-Latin-script languages would help assess the generality of these findings and inform multilingual AI governance; we focus on Bangla as a representative case study to enable in-depth analysis within limited computational and experimental resources.

\section{Conclusion}
This work presents the first systematic analysis of text watermarking robustness for Bangla LLMs under cross-lingual round-trip translation attacks. 
Through extensive empirical evaluation, we show that while state-of-the-art token-level watermarking methods remain effective under benign conditions, their detectability collapses under RTT due to fundamental disruptions in token distributions \cite{kirchenbauer2023watermark,kuditipudi2023robust,he2024can}. 
To address this limitation, we introduce a layered watermarking strategy that reinforces embedding-time watermarks with a post-generation statistical signal. 
Experimental results demonstrate that this double-layer design significantly improves watermark survivability under RTT, albeit at the cost of controlled semantic degradation \cite{lau2024waterfall}. 
These findings underscore the necessity of multilingual robustness evaluation and highlight layered watermarking as a practical, training-free approach for strengthening watermark detection in low-resource language settings. 
This study provides practical insights for the resilient and responsible implementation of LLM watermarking in low-resource languages by highlighting potential failure modes and proposing effective mitigation strategies.

\section{Limitations}

While our experiments provide valuable insights into watermarking robustness, several limitations remain. First, the evaluation focuses on a single RTT pathway (Bangla$\rightarrow$English$\rightarrow$Bangla), which, while realistic and widely used, does not capture the full space of possible cross-lingual or multi-hop paraphrasing attacks \cite{he2024can}; future work could investigate additional RTT pathways and multilingual transformations to better understand cross-lingual watermark robustness. Second, the double-layer watermarking experiments are restricted to shorter generation lengths due to computational constraints, limiting analysis of how layered robustness scales with longer text; exploring longer sequences and scalable watermarking strategies could provide more in-depth insights under realistic generation conditions. Third, although ROUGE and perplexity provide useful proxies for semantic preservation and fluency, they may not fully capture subtle meaning drift introduced by layered perturbations, particularly in morphologically rich languages such as Bangla \cite{hasan2020not}; incorporating human evaluation or more fine-grained semantic similarity metrics could address this gap. Finally, the study evaluates a fixed set of watermarking algorithms and does not explore adaptive or learned combinations of layers, which could further improve robustness but introduce additional system complexity; developing adaptive or hybrid watermarking strategies represents a promising direction for enhancing resilience against other complex attacks.
\section{Ethical Considerations and Broader Impact}

This work examines text watermarking methods for Bangla LLM generation to support authorship attribution, provenance verification, and misuse detection. While watermarking can contribute to responsible AI deployment, it also introduces ethical challenges related to interpretability, fairness, and potential misuse, particularly in low-resource language contexts.

\paragraph{Responsible Interpretation.}
Watermark detection is inherently statistical and probabilistic rather than definitive. As such, detected watermark signals should not be treated as conclusive evidence of AI authorship in isolation. Overreliance on watermark-based attribution may lead to false positives or misclassification of legitimate human-authored text. Consistent with prior responsible NLP guidance, we emphasize that watermarking should complement broader governance mechanisms, including transparency, contextual analysis, and human oversight, rather than serving as a standalone enforcement tool \citep{venkatakrishnan2023watermarking}.

\paragraph{Dual-Use Risks.}
Text watermarking constitutes a dual-use technology. While it can help mitigate large-scale misuse of generative models, it could also be misapplied for censorship, surveillance, or non-consensual monitoring of content creators. To reduce such risks, our approach operates under strict black-box assumptions, does not encode user-identifying information, and embeds only statistical provenance signals. These design choices align with prior recommendations to minimize privacy and misuse risks in watermarking systems \citep{kirchenbauer2023watermark,kuditipudi2023robust}.

\paragraph{Fairness in Low-Resource Languages.}
A central motivation of this work is addressing the lack of robustness evaluations for low-resource and non-Latin-script languages. English-centric watermarking assumptions often fail in morphologically rich languages such as Bangla, potentially resulting in unequal reliability across linguistic communities. By systematically evaluating watermarking under cross-lingual round-trip translation (RTT) attacks in Bangla, this work contributes toward more equitable multilingual AI governance. However, imperfect robustness may still disproportionately affect low-resource language users if watermark-based policies are applied asymmetrically \citep{venkatakrishnan2023watermarking}.

\paragraph{Quality--Robustness Trade-off.}
The proposed layered watermarking strategy improves robustness under RTT at the cost of controlled semantic perturbations. While acceptable for forensic attribution and provenance verification, such degradation may be unsuitable for high-stakes or user-facing applications. We therefore recommend selective deployment based on application context and threat models, accompanied by clear disclosure regarding the presence and purpose of watermarking \citep{lau2024waterfall}.

Overall, this work aims to advance responsible multilingual LLM deployment by identifying robustness limitations of existing watermarking techniques and proposing practical, training-free mitigations for low-resource languages.

\bibliography{citations}
\newpage
\appendix
\section{Appendix}

\subsection{Overview of Experimental Pipeline}
\label{app:overview}

This appendix provides implementation-level details to support reproducibility of our layered watermarking experiments, while omitting full source code due to space constraints. 
Our experimental pipeline consists of three sequential stages: (i) primary watermark generation using baseline methods (EXP and KGW), (ii) secondary paraphrase-based watermarking using the Waterfall framework, and (iii) detection and robustness evaluation under round-trip translation (RTT) attacks. 
All experiments are conducted exclusively on Bangla text, using Bangla-adapted prompts, models, and evaluation procedures.

\subsection{Primary Watermarking Algorithms and Configuration}
\label{app:primary}

We employ two established watermarking methods as baselines: EXP and KGW. Both algorithms are implemented following their original formulations, without modification to their detection statistics or theoretical assumptions.

\paragraph{EXP.}
EXP watermarking is implemented using exponential sampling, where the random seed is deterministically derived from the preceding token prefix. Detection is performed using a Gamma survival function over accumulated token-wise scores.

\paragraph{KGW.}
KGW watermarking biases token generation toward a dynamically constructed greenlist and detects watermarks via a z-score threshold on green-token counts.Table~\ref{tab:exp_kgw_params} summarizes the exact hyperparameters used across all experiments.

\begin{table}[h]
\centering
\small
\caption{Primary watermarking hyperparameters (held constant across all experiments).}
\label{tab:exp_kgw_params}
\begin{tabular}{lcc}
\toprule
\textbf{Parameter} & \textbf{EXP} & \textbf{KGW} \\
\midrule
Prefix length & 4 & 1 \\
Hash key & 15485863 & 15485863 \\
Sequence length & 200 & -- \\
Detection threshold & $10^{-4}$ & $z > 4.0$ \\
$\gamma$ (green-list ratio) & -- & 0.5 \\
$\delta$ (bias strength) & -- & 2.0 \\
\bottomrule
\end{tabular}
\end{table}

\subsection{Detection Score Computation for KGW and EXP}
\label{sec:appendix_detection_scores}

This section describes how detection scores are computed for the KGW and Exponential Sampling (EXP) watermarking methods \cite{kirchenbauer2023watermark,kuditipudi2023robust}. Although both approaches yield scalar detection statistics, the underlying definitions and statistical interpretations differ substantially.

\paragraph{KGW Detection Score.}
KGW watermarking partitions the vocabulary at each decoding step into a pseudorandom \emph{green list} and \emph{red list} using a secret hash key. During generation, sampling is biased toward green-list tokens. Let $T$ denote the total number of generated tokens and $G$ the number of tokens belonging to the green list. Under the null hypothesis of unwatermarked text, $G$ follows a binomial distribution with expected value $\mathbb{E}[G] = \gamma T$ and variance $\mathrm{Var}(G) = \gamma(1-\gamma)T$, where $\gamma$ is the green-list ratio \cite{kirchenbauer2023watermark}.

The KGW detection score is computed as a Z-statistic:
\[
z = \frac{G - \gamma T}{\sqrt{\gamma(1-\gamma)T}}.
\]
A watermark is detected if the resulting Z-score exceeds a predefined threshold (e.g., $z > 4.0$).

\paragraph{EXP Detection Score.}
EXP watermarking embeds watermarks through distortion-free exponential sampling, without explicitly biasing token probabilities. Each token has a score $s_t$, accumulated across the sequence to produce:
\[
S = \sum_{t=1}^{T} s_t,
\]
where $s_t$ denotes the exponential sampling score at position $t$ \cite{kuditipudi2023robust}.

Under the null hypothesis of unwatermarked text, $S$ follows a Gamma distribution. Detection is performed by computing a $p$-value using the Gamma survival function, with a watermark declared present if $p < 10^{-4}$.

\paragraph{Key Differences.}
KGW detection relies on discrete token-count statistics and hypothesis testing over green-token frequencies, whereas EXP detection is likelihood-based and operates on continuous accumulated scores. Consequently, KGW scores are highly sensitive to token substitution, while EXP scores degrade more gradually under distributional shifts such as cross-lingual paraphrasing \cite{kuditipudi2023robust}

\subsection{Secondary Watermarking via Paraphrase Selection}
\label{app:secondary}

To improve robustness against paraphrasing-based attacks, we apply a secondary watermarking layer using the Waterfall framework. Given a primary-watermarked Bangla text, the model generates multiple paraphrase candidates under a watermark-aware decoding process.

Each candidate is evaluated using two criteria:
(i) a watermark verification score derived from the Waterfall detector, and
(ii) semantic similarity to the original text, measured using a sentence embedding model.
The final output is selected by maximizing a weighted combination of semantic similarity and watermark strength.

\begin{algorithm}[t]
\caption{Layered Watermarking via Semantic-Aware Paraphrase Selection}
\label{alg:layered}
\KwIn{Primary-watermarked Bangla text $T_o$}
\KwOut{Layered watermarked text $T_w$}

\KwData{Secondary watermarking function $\mathcal{W}$, 
semantic similarity model $\mathcal{S}$, 
weight parameter $\lambda$}

\SetKwProg{Fn}{Function}{:}{}

\Fn{LayeredWatermark($T_o$)}{
1. Generate a set of candidate paraphrases $\mathcal{C} = \{\mathcal{W}(T_o)_i\}_{i=1}^N$\;
2. \ForEach{$T_i \in \mathcal{C}$}{
3. \quad Compute semantic similarity $s_i = \mathcal{S}(T_o, T_i)$\;
4. \quad Compute watermark verification score $q_i$\;
}
5. Select $T_w = \arg\max_{T_i \in \mathcal{C}} \left( \lambda \cdot s_i + q_i \right)$\;
6. \Return $T_w$\;
}
\end{algorithm}

This selection mechanism ensures that watermark robustness is improved without compromising semantic fidelity.

\subsection{Bangla-Specific Adaptations}
\label{app:bangla}

All prompts, generated outputs, and evaluation steps are conducted in Bangla. 
Paraphrasing prompts explicitly restrict the model to Bangla-only generation and disallow explanatory or summarizing text.
While watermarking algorithms are language-agnostic in principle, tokenization behavior, paraphrase diversity, and RTT robustness differ substantially for low-resource, non-Latin scripts.
Our study therefore focuses on Bangla as a representative low-resource language, rather than assuming transferability from English-centric evaluations.

\subsection{Round-Trip Translation (RTT) Evaluation}
\label{app:rtt}

Robustness is evaluated using round-trip translation attacks, where Bangla text is translated to an intermediate language and back to Bangla.
We report detection accuracy before and after RTT for both single-layer and layered watermarking.
RTT protocols and translation paths are held fixed across methods to ensure fair comparison.

\subsection{Data Artifacts and CSV Schema}
\label{app:data}

Intermediate and final outputs are stored in structured CSV files to enable modular evaluation across experimental stages.
Each CSV contains the original prompt, primary-watermarked output, secondary-watermarked output (if applicable), detection statistics, and semantic similarity scores.

\begin{table}[h]
\centering
\small
\caption{CSV schema used across experiments.}
\label{tab:csv_schema}
\begin{tabular}{lp{0.6\linewidth}}
\toprule
\textbf{Column} & \textbf{Description} \\
\midrule
input & Original Bangla input text \\
primary\_wm & Output after primary watermarking (EXP or KGW) \\
secondary\_wm & Output after secondary Waterfall watermarking \\
q\_score & Waterfall watermark verification score \\
sts\_score & Semantic similarity with original text \\
rtt\_detected & Detection result after RTT attack \\
\bottomrule
\end{tabular}
\end{table}

\subsection{Implementation Notes and Reproducibility}
\label{app:repro}

All experiments are implemented using Python and PyTorch, with Hugging Face \texttt{transformers} for model inference.
Due to computational constraints, models are loaded using 8-bit quantization where applicable.
We retain identical decoding and detection settings across all comparative experiments.
Full source code and configuration files will be released publicly upon publication.

\end{document}